\documentstyle[aclap]{article}

\title{Tagset Design and Inflected Languages}
\author{David Elworthy\\
Sharp Laboratories of Europe Ltd\\
Edmund Halley Road\\
Oxford Science Park\\
Oxford OX4 4GA\\
{\tt dahe@sharp.co.uk}}
\date{17 January 1995}

\begin{document}
\maketitle
\begin{abstract}
An experiment designed to explore the relationship between tagging accuracy
and the nature of the tagset is described, using corpora in English, French
and Swedish. In particular, the question of internal versus external criteria
for tagset design is considered, with the general conclusion that external
(linguistic) criteria should be followed. Some problems associated with
tagging unknown words in inflected languages are briefly considered.
\end{abstract}
\bibliographystyle{acl}

\section{Tagset Design}
Tagging by means of a Hidden Markov Model (HMM) is  widely recognised as an
effective technique for assigning parts of speech to a corpus in a robust and
efficient manner. An attractive feature of the technique is that the
algorithm itself is independent of the (natural) language to which it is
applied. All of the ``knowledge engineering'' is localised in the choice of
tagset and the method of training. Typically, training makes use of a manually
tagged corpus, or an untagged corpus with some initial bootstrapping
probabilities. Some attention has been given to how to make such techniques
effective; for example Cutting et al.~(1992) \nocite{Cutting:pos} suggest
ways of training trigram taggers, and Merialdo~(1994)
\nocite{Merialdo:tagging} and Elworthy~(1994) \nocite{Elworthy:help} consider
the amount and quality of the seeding data needed to construct an accurate
tagger.

In training a tagger for a given language, a major part of the knowledge
engineering required can therefore be localised in the choice of the tagset.
The design of an appropriate tagset is subject to both external and internal
criteria. The external criterion is that the tagset must be capable of making
the linguistic (for example, syntactic or morphological) distinctions required
in the output corpora. Tagsets used in the past have included varying amounts
of detail. For example, the Penn treebank tagset \cite{Marcus:penn} omits a
number of the distinctions which are made in the LOB and Brown tagsets on
which it is based \cite{Garside:LAE,Francis:brown} in cases where the surface
form of the words allows the distinctions to be recovered if they are
needed. Thus, the auxiliary verbs {\em be}, {\em do} and {\em have} have the
same tags as other verbs in Penn, but are each separated out in the LOB
tagset.

A second design criterion on tagsets is the internal one of making the tagging
as effective as possible. As an example, one of the most common errors made by
taggers with the LOB and Brown tagsets is mistagging a word as a subordinating
conjunction (CS) rather than a preposition (IN), or vice-versa
\cite{Macklovitch:falter}. A higher level of syntactic analysis indicating the
phrasal structure would be required to predict which tag is correct, and this
information is not available to fixed-context taggers.  The Penn treebank
therefore uses a single tag for both cases, leaving the resolution -- if
required -- to some other process. Similarly, most tagsets do not distinguish
transitive and intransitive verbs, since taggers which use a context of only
two or three words will generally not be able to make the right
predictions. Distinctions of this sort are usually found only in corpora
such as Susanne which are parsed as well as tagged.

The problem of tagset design becomes particularly important for highly
inflected languages, such as Greek or Hungarian. If all of the syntactic
variations which are realised in the inflectional system were represented in
the tagset, there would be a huge number of tags, and it would be practically
impossible to implement or train a simple tagger. Note in passing that this
may not as serious a problem as it first appears. If the language is very
highly inflected, it may be be possible to do all (or a large part) of the
work of a tagger with a word-by-word morphological analysis instead.
Nevertheless, there are many languages which have enough ambiguity that
tagging is useful, but a rich enough tagset that the criteria on which it is
designed must be given careful consideration.

In this paper, I report two experiments which address the {\em internal}
design criterion, by looking at how tagging accuracy varies as the tagset is
modified, in English, French and Swedish. Although the choice of language was
dictated by the corpora which were available, they represent three different
degrees of complexity in their inflectional systems. English has a very limited
system, marking little more than plurality on nouns and a restricted range of
verb properties. French has a little more complexity, with gender, number and
person marked, while Swedish has more detailed marking for gender, number,
definiteness and case.  As a subsidiary issue, we will also look at how the
tagger performs on unknown words, i.e. ones not seen in the training data. The
usual approach here is to hypothesise all tags in the tagset for an unknown
word, other than ones where all the words that may have the tag can be
enumerated in advance (closed class tags). HMM taggers often perform poorly on
unknown words.

Alternative tagsets were derived by taking the initial tagset for each corpus
(from manual tagging of the corpus) and condensing sets of tags which
represent a grammatical distinction such as gender into single tags. The
changes were then applied to the training corpus. This allows us to
effectively produce a corpus tagged according to a different scheme without
having to manually re-tag the corpus. The changes in the tagsets were
motivated purely by grammatical considerations, and did not take the errors
actually observed into account. In general what we will look at in the results
is how the tagging accuracy changes as the {\em size} of the tagset
changes. This is a deliberately naive approach, and it is adopted with the
goal of continuing in the relatively ``knowledge-free'' tradition of work in
HMM tagging. The aim of the experiment is to determine, crudely, whether a
bigger tagset is better than a smaller one, or whether external criteria
requiring human intervention should be used to choose the best tagset. The
results for the three languages turn out to be quite different, and the
general conclusion (which is the overall contribution of the paper) will be
that the external criterion should be the one to dominate tagset design: there
is a limit to how knowledge-free we can be.

As a preliminary to this work, note that it is hard to reason about the
effect of changing the tagset. It can be argued that a smaller tagset should
improve tagging accuracy, since it puts less of a burden on the tagger to make
fine distinctions. In information-theoretic terms, the number of decisions
required is smaller, and hence the tagger need contribute less information to
make the decisions.  A smaller tagset may also mean that more words have only
one possible tag and so can be handled trivially. Conversely, more detail in
the tagset may help the tagger when the properties of two adjacent words give
support to the choice of tag for both of them; that is, the transitions
between tags contribute the information the tagger needs. For example, if
determiners and nouns are marked for number, then the tagger can effectively
model agreement in simple noun phrases, by having a higher probability for a
singular determiner followed by a singular noun that it does for a singular
determiner followed by a plural noun. Theory on its own does not help much in
deciding which point of view should dominate.

\section{The experiments}

\subsection{Design of the experiments}
Two experiments were conducted on three corpora: 300k words of Swedish text
from the ECI Multilingual CD-ROM, and 100k words each of English and French
from a corpus of International Telecommunications Union text\footnote{The
English and French corpora were kindly supplied to us by Tony McEnery,
and are translation-equivalent. See McEnery et al.~(1994)
\nocite{McEnery:parallel} for details.}. In the first experiment the whole of
each corpus was used to train the model, and a small sample from the same text
was used as test data. For the second experiment, 95\% of the corpus was used
in training and the remainder in testing. The importance of the second test is
that it includes unknown words, which are difficult to tag. The tagsets were
progressively modified, by textually substituting simplified tags for the
original ones and e e-running the training and test procedures using the
modified corpora. The changes to the tagset are listed below. In the results
that follow, we will identify tagset that include a given distinction with an
uppercase letter and ones that do not with a lowercase letter; for example
{\em G} for a tagset that marks gender, and {\em g} for one that does not.
\begin{description}
\item[Swedish] The changes made were entirely based on inflections.
\begin{description}
\item[G] Gender: masculine, neuter, common gender (``UTR'' in the tagset).
\item[N] Number: singular, plural.
\item[D] Definiteness: definite, indefinite.
\item[C] Case: nominative, genitive.
\end{description}
\item[French] The changes other than {\em V} were based on inflections.
\begin{description}
\item[G] Gender: masculine, feminine.
\item[N] Number: singular, plural.
\item[P] Person: identified as 1st to 6th in the tagset.
\item[V] Verbs: treat {\em avoir} and {\em etre} as being the same as
any other verb.
\end{description}
\item[English] The changes here are more varied than for the other
languages, and generally consisted of removing some of the finer
subdivisions of the major classes. The grouping of some of these changes
is admittedly a little {\em ad hoc}, and was intended to give a good
distribution of tagset sizes; not all combinations were tried.
\begin{description}
\item[C] Reduce specific conjunction classes to a common class,
and simplify one adjective class.
\item[A] Simplify noun and adverb classes.
\item[P] Simplify pronoun classes.
\item[N] Number: all singular/plural distinctions removed.
\item[V] Use the same class for {\em have}, {\em do} and {\em be} as for
other verbs.
\end{description}
\end{description}
The sizes of the resulting tagsets and the degree of ambiguity in the corpora
which resulted appear below. Accuracy figures quoted here are for ambiguous
and unknown words only, and therefore factor out effects due to the varying
degree of ambiguity as the tagset changes. In fact, this is a rather
approximate way of accounting for ambiguity, since it does not take the length
of ambiguous sequences into account, and the accuracy is likely to deteriorate
more on long sequences of ambiguous words than on short ones.

The tests were run using Good-Turing correction to the probability estimates;
that is, rather than estimating the probability of the transition from a tag
$i$ to a tag $j$ as the count of transition from $i$ to $j$ in the training
corpus divided by the total frequency of tag $i$, one was added to the count
of all transitions, and the total tag frequencies adjusted
correspondingly. The purpose in using this correction is to correct for
corpora which might not provide enough training data. On the largest tagsets,
the correction was found to give a very slight reduction in the accuracy for
Swedish, and to improve the French and English accuracies by about 1.5\%,
suggesting that it is indeed needed.

\subsection{Results}

The first experiment, with no unknown words, gave accuracies on ambiguous
words of 91--93\% for Swedish, 94--97\% for French and 85--90\% for
English. The results for English are surprisingly low (for example, on the
Penn treebank, the tagger gives an accuracy of 95--96\%), and may be due to
long sequences of ambiguous words. The results appear in
table~\ref{tab1}. The figures include the degree of
ambiguity, that is, the number of words in the corpus for which more than one
tag was hypothesised. The accuracy is plotted against the size of the tagset
in figures~\ref{fig-swe1}--\ref{fig-eng1}, where the numbers on the points
correspond to the index of tagsets listed. Summarising the patterns:
\begin{description}
\item[Swedish] Larger tagset generally gives higher accuracy. The results are
quite widely spread.
\item[French] Clustered, with an accuracy on all tagsets which do not mark
gender of around 96\%--96.5\%; when gender is marked 94\%--94.5\%.
\item[English] Larger tagset tends to give larger accuracy, though with less
of a spread than for Swedish.
\end{description}
The sizes of the tagsets ranged from approximately 80--200 tags for Swedish,
35--90 for French, and 70--160 for English. As discussed above, it is not
clear what would happen with larger tagsets, but some experiments based on the
Susanne corpus and using tagsets ranging from 236 to 425 tags suggest that the
trend to higher accuracy continues with even bigger tagsets.
\begin{table*}[p]
\begin{center}
\caption{Results for test with no unknown words}
\begin{tabular}{|c|c|c|c|c|c|} \hline
Language & Index & Tagset & Size of & Degree of      & Ambiguous word \\
         &       &        & tagset  & ambiguity (\%) & accuracy (\%) \\ \hline
Swedish  & 1     & GNDC & 194 & 41.57 & 92.02\\
         & 2     & GnDC & 170 & 39.29 & 92.23\\
         & 3     & GNDc & 167 & 41.49 & 91.92\\
         & 4     & GNdC & 162 & 41.45 & 91.67\\
         & 5     & gNDC & 152 & 41.54 & 91.88\\
         & 6     & GnDc & 147 & 39.21 & 92.04\\
         & 7     & GndC & 141 & 37.43 & 91.86\\
         & 8     & GNdc & 140 & 41.37 & 91.63\\
         & 9     & gNDc & 134 & 41.47 & 91.82\\
         & 10    & gNdC & 126 & 41.42 & 91.34\\
         & 11    & Gndc & 123 & 37.36 & 91.74\\
         & 12    & gnDC & 121 & 39.18 & 91.35\\
         & 13    & gNdc & 113 & 41.34 & 91.29\\
         & 14    & gnDc & 105 & 39.11 & 91.28\\
         & 15    & gndC & 96 & 37.32 & 91.56\\
         & 16    & gndc & 86 & 37.25 & 91.52\\\hline
French   & 1     & GNPV & 87 & 49.77 & 94.43\\
         & 2     & GNPv & 80 & 49.75 & 94.35\\
         & 3     & GNpV & 76 & 49.77 & 94.31\\
         & 4     & GNpv & 74 & 49.75 & 94.39\\
         & 5     & GnPV & 64 & 49.49 & 94.28\\
         & 6     & gNPV & 62 & 47.48 & 96.34\\
         & 7     & GnPv & 57 & 49.47 & 94.36\\
         & 8     & gNPv & 55 & 47.64 & 96.22\\
         & 9     & GnpV & 53 & 49.49 & 94.25\\
         & 10    & gNpV & 51 & 47.48 & 96.14\\
         & 11    & Gnpv & 51 & 49.47 & 94.36\\
         & 12    & gnPV & 49 & 47.03 & 96.34\\
         & 13    & gNpv & 49 & 47.46 & 96.10\\
         & 14    & gnPv & 42 & 47.01 & 96.30\\
         & 15    & gnpV & 38 & 47.03 & 96.30\\
         & 16    & gnpv & 36 & 47.01 & 96.34\\\hline
English  & 1     & CAPNV & 153 & 47.95 & 89.56\\
         & 2     & CApNV & 150 & 47.47 & 89.27\\
         & 3     & cAPNV & 145 & 47.91 & 89.50\\
         & 4     & CAPNv & 140 & 47.95 & 89.33\\
         & 5     & CAPnV & 137 & 47.95 & 89.20\\
         & 6     & CaPNV & 129 & 47.95 & 89.20\\
         & 7     & CAPnv & 124 & 47.95 & 89.01\\
         & 8     & capNV & 119 & 47.43 & 88.94\\
         & 9     & capnV & 108 & 47.13 & 88.45\\
         & 10    & capNv & 106 & 47.43 & 88.48\\
         & 11    & capnv & 95 & 47.13 & 85.42\\\hline
\end{tabular}
\label{tab1}
\end{center}
\end{table*}
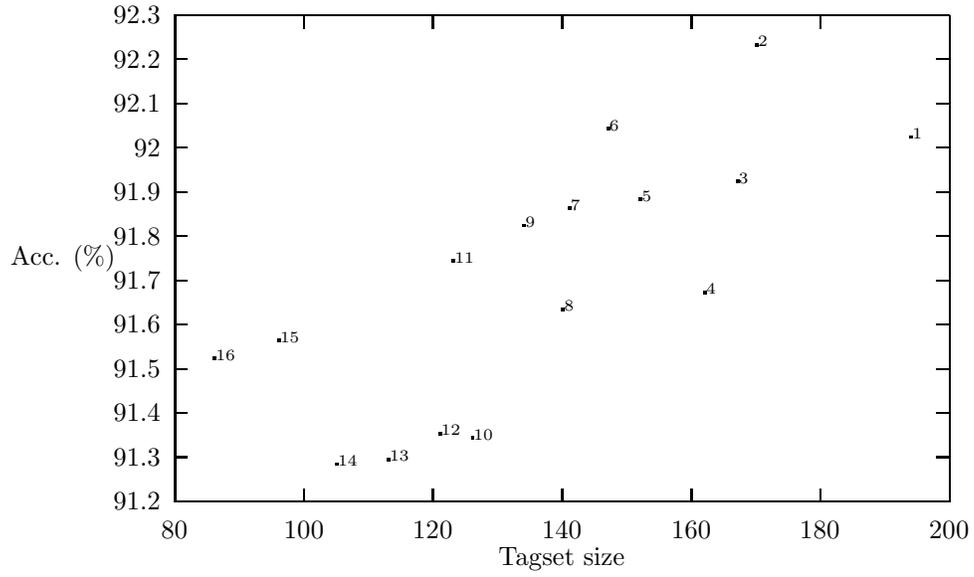
\begin{figure*}
\begin{center}
\item[]
\setlength{\unitlength}{0.240900pt}
\ifx\plotpoint\undefined\newsavebox{\plotpoint}\fi
\begin{picture}(1500,900)(0,0)
\font\gnuplot=cmr10 at 10pt
\gnuplot
\sbox{\plotpoint}{\rule[-0.200pt]{0.400pt}{0.400pt}}%
\put(220.0,113.0){\rule[-0.200pt]{4.818pt}{0.400pt}}
\put(198,113){\makebox(0,0)[r]{91.2}}
\put(1416.0,113.0){\rule[-0.200pt]{4.818pt}{0.400pt}}
\put(220.0,182.0){\rule[-0.200pt]{4.818pt}{0.400pt}}
\put(198,182){\makebox(0,0)[r]{91.3}}
\put(1416.0,182.0){\rule[-0.200pt]{4.818pt}{0.400pt}}
\put(220.0,252.0){\rule[-0.200pt]{4.818pt}{0.400pt}}
\put(198,252){\makebox(0,0)[r]{91.4}}
\put(1416.0,252.0){\rule[-0.200pt]{4.818pt}{0.400pt}}
\put(220.0,321.0){\rule[-0.200pt]{4.818pt}{0.400pt}}
\put(198,321){\makebox(0,0)[r]{91.5}}
\put(1416.0,321.0){\rule[-0.200pt]{4.818pt}{0.400pt}}
\put(220.0,391.0){\rule[-0.200pt]{4.818pt}{0.400pt}}
\put(198,391){\makebox(0,0)[r]{91.6}}
\put(1416.0,391.0){\rule[-0.200pt]{4.818pt}{0.400pt}}
\put(220.0,460.0){\rule[-0.200pt]{4.818pt}{0.400pt}}
\put(198,460){\makebox(0,0)[r]{91.7}}
\put(1416.0,460.0){\rule[-0.200pt]{4.818pt}{0.400pt}}
\put(220.0,530.0){\rule[-0.200pt]{4.818pt}{0.400pt}}
\put(198,530){\makebox(0,0)[r]{91.8}}
\put(1416.0,530.0){\rule[-0.200pt]{4.818pt}{0.400pt}}
\put(220.0,599.0){\rule[-0.200pt]{4.818pt}{0.400pt}}
\put(198,599){\makebox(0,0)[r]{91.9}}
\put(1416.0,599.0){\rule[-0.200pt]{4.818pt}{0.400pt}}
\put(220.0,669.0){\rule[-0.200pt]{4.818pt}{0.400pt}}
\put(198,669){\makebox(0,0)[r]{92}}
\put(1416.0,669.0){\rule[-0.200pt]{4.818pt}{0.400pt}}
\put(220.0,738.0){\rule[-0.200pt]{4.818pt}{0.400pt}}
\put(198,738){\makebox(0,0)[r]{92.1}}
\put(1416.0,738.0){\rule[-0.200pt]{4.818pt}{0.400pt}}
\put(220.0,808.0){\rule[-0.200pt]{4.818pt}{0.400pt}}
\put(198,808){\makebox(0,0)[r]{92.2}}
\put(1416.0,808.0){\rule[-0.200pt]{4.818pt}{0.400pt}}
\put(220.0,877.0){\rule[-0.200pt]{4.818pt}{0.400pt}}
\put(198,877){\makebox(0,0)[r]{92.3}}
\put(1416.0,877.0){\rule[-0.200pt]{4.818pt}{0.400pt}}
\put(220.0,113.0){\rule[-0.200pt]{0.400pt}{4.818pt}}
\put(220,68){\makebox(0,0){80}}
\put(220.0,857.0){\rule[-0.200pt]{0.400pt}{4.818pt}}
\put(423.0,113.0){\rule[-0.200pt]{0.400pt}{4.818pt}}
\put(423,68){\makebox(0,0){100}}
\put(423.0,857.0){\rule[-0.200pt]{0.400pt}{4.818pt}}
\put(625.0,113.0){\rule[-0.200pt]{0.400pt}{4.818pt}}
\put(625,68){\makebox(0,0){120}}
\put(625.0,857.0){\rule[-0.200pt]{0.400pt}{4.818pt}}
\put(828.0,113.0){\rule[-0.200pt]{0.400pt}{4.818pt}}
\put(828,68){\makebox(0,0){140}}
\put(828.0,857.0){\rule[-0.200pt]{0.400pt}{4.818pt}}
\put(1031.0,113.0){\rule[-0.200pt]{0.400pt}{4.818pt}}
\put(1031,68){\makebox(0,0){160}}
\put(1031.0,857.0){\rule[-0.200pt]{0.400pt}{4.818pt}}
\put(1233.0,113.0){\rule[-0.200pt]{0.400pt}{4.818pt}}
\put(1233,68){\makebox(0,0){180}}
\put(1233.0,857.0){\rule[-0.200pt]{0.400pt}{4.818pt}}
\put(1436.0,113.0){\rule[-0.200pt]{0.400pt}{4.818pt}}
\put(1436,68){\makebox(0,0){200}}
\put(1436.0,857.0){\rule[-0.200pt]{0.400pt}{4.818pt}}
\put(220.0,113.0){\rule[-0.200pt]{292.934pt}{0.400pt}}
\put(1436.0,113.0){\rule[-0.200pt]{0.400pt}{184.048pt}}
\put(220.0,877.0){\rule[-0.200pt]{292.934pt}{0.400pt}}
\put(45,495){\makebox(0,0){Acc. (\%)}}
\put(828,23){\makebox(0,0){Tagset size}}
\put(220.0,113.0){\rule[-0.200pt]{0.400pt}{184.048pt}}
\put(1375,683){\rule{1pt}{1pt}{\tiny 1}}
\put(1132,828){\rule{1pt}{1pt}{\tiny 2}}
\put(1102,613){\rule{1pt}{1pt}{\tiny 3}}
\put(1051,439){\rule{1pt}{1pt}{\tiny 4}}
\put(950,585){\rule{1pt}{1pt}{\tiny 5}}
\put(899,696){\rule{1pt}{1pt}{\tiny 6}}
\put(838,571){\rule{1pt}{1pt}{\tiny 7}}
\put(828,412){\rule{1pt}{1pt}{\tiny 8}}
\put(767,544){\rule{1pt}{1pt}{\tiny 9}}
\put(686,210){\rule{1pt}{1pt}{\tiny 10}}
\put(656,488){\rule{1pt}{1pt}{\tiny 11}}
\put(635,217){\rule{1pt}{1pt}{\tiny 12}}
\put(554,176){\rule{1pt}{1pt}{\tiny 13}}
\put(473,169){\rule{1pt}{1pt}{\tiny 14}}
\put(382,363){\rule{1pt}{1pt}{\tiny 15}}
\put(281,335){\rule{1pt}{1pt}{\tiny 16}}
\end{picture}
\caption{Results for Swedish (no unknown words)}
\label{fig-swe1}
\end{center}
\end{figure*}
\begin{figure*}
\begin{center}
\item[]
\setlength{\unitlength}{0.240900pt}
\ifx\plotpoint\undefined\newsavebox{\plotpoint}\fi
\begin{picture}(1500,900)(0,0)
\font\gnuplot=cmr10 at 10pt
\gnuplot
\sbox{\plotpoint}{\rule[-0.200pt]{0.400pt}{0.400pt}}%
\put(220.0,113.0){\rule[-0.200pt]{4.818pt}{0.400pt}}
\put(198,113){\makebox(0,0)[r]{94}}
\put(1416.0,113.0){\rule[-0.200pt]{4.818pt}{0.400pt}}
\put(220.0,266.0){\rule[-0.200pt]{4.818pt}{0.400pt}}
\put(198,266){\makebox(0,0)[r]{94.5}}
\put(1416.0,266.0){\rule[-0.200pt]{4.818pt}{0.400pt}}
\put(220.0,419.0){\rule[-0.200pt]{4.818pt}{0.400pt}}
\put(198,419){\makebox(0,0)[r]{95}}
\put(1416.0,419.0){\rule[-0.200pt]{4.818pt}{0.400pt}}
\put(220.0,571.0){\rule[-0.200pt]{4.818pt}{0.400pt}}
\put(198,571){\makebox(0,0)[r]{95.5}}
\put(1416.0,571.0){\rule[-0.200pt]{4.818pt}{0.400pt}}
\put(220.0,724.0){\rule[-0.200pt]{4.818pt}{0.400pt}}
\put(198,724){\makebox(0,0)[r]{96}}
\put(1416.0,724.0){\rule[-0.200pt]{4.818pt}{0.400pt}}
\put(220.0,877.0){\rule[-0.200pt]{4.818pt}{0.400pt}}
\put(198,877){\makebox(0,0)[r]{96.5}}
\put(1416.0,877.0){\rule[-0.200pt]{4.818pt}{0.400pt}}
\put(220.0,113.0){\rule[-0.200pt]{0.400pt}{4.818pt}}
\put(220,68){\makebox(0,0){30}}
\put(220.0,857.0){\rule[-0.200pt]{0.400pt}{4.818pt}}
\put(423.0,113.0){\rule[-0.200pt]{0.400pt}{4.818pt}}
\put(423,68){\makebox(0,0){40}}
\put(423.0,857.0){\rule[-0.200pt]{0.400pt}{4.818pt}}
\put(625.0,113.0){\rule[-0.200pt]{0.400pt}{4.818pt}}
\put(625,68){\makebox(0,0){50}}
\put(625.0,857.0){\rule[-0.200pt]{0.400pt}{4.818pt}}
\put(828.0,113.0){\rule[-0.200pt]{0.400pt}{4.818pt}}
\put(828,68){\makebox(0,0){60}}
\put(828.0,857.0){\rule[-0.200pt]{0.400pt}{4.818pt}}
\put(1031.0,113.0){\rule[-0.200pt]{0.400pt}{4.818pt}}
\put(1031,68){\makebox(0,0){70}}
\put(1031.0,857.0){\rule[-0.200pt]{0.400pt}{4.818pt}}
\put(1233.0,113.0){\rule[-0.200pt]{0.400pt}{4.818pt}}
\put(1233,68){\makebox(0,0){80}}
\put(1233.0,857.0){\rule[-0.200pt]{0.400pt}{4.818pt}}
\put(1436.0,113.0){\rule[-0.200pt]{0.400pt}{4.818pt}}
\put(1436,68){\makebox(0,0){90}}
\put(1436.0,857.0){\rule[-0.200pt]{0.400pt}{4.818pt}}
\put(220.0,113.0){\rule[-0.200pt]{292.934pt}{0.400pt}}
\put(1436.0,113.0){\rule[-0.200pt]{0.400pt}{184.048pt}}
\put(220.0,877.0){\rule[-0.200pt]{292.934pt}{0.400pt}}
\put(45,495){\makebox(0,0){Acc. (\%)}}
\put(828,23){\makebox(0,0){Tagset size}}
\put(220.0,113.0){\rule[-0.200pt]{0.400pt}{184.048pt}}
\put(1375,244){\rule{1pt}{1pt}{\tiny 1}}
\put(1233,220){\rule{1pt}{1pt}{\tiny 2}}
\put(1152,208){\rule{1pt}{1pt}{\tiny 3}}
\put(1112,232){\rule{1pt}{1pt}{\tiny 4}}
\put(909,199){\rule{1pt}{1pt}{\tiny 5}}
\put(869,828){\rule{1pt}{1pt}{\tiny 6}}
\put(767,223){\rule{1pt}{1pt}{\tiny 7}}
\put(727,791){\rule{1pt}{1pt}{\tiny 8}}
\put(686,189){\rule{1pt}{1pt}{\tiny 9}}
\put(646,767){\rule{1pt}{1pt}{\tiny 10}}
\put(646,223){\rule{1pt}{1pt}{\tiny 11}}
\put(605,828){\rule{1pt}{1pt}{\tiny 12}}
\put(605,755){\rule{1pt}{1pt}{\tiny 13}}
\put(463,816){\rule{1pt}{1pt}{\tiny 14}}
\put(382,816){\rule{1pt}{1pt}{\tiny 15}}
\put(342,828){\rule{1pt}{1pt}{\tiny 16}}
\end{picture}
\caption{Results for French (no unknown words)}
\label{fig-fre1}
\end{center}
\end{figure*}
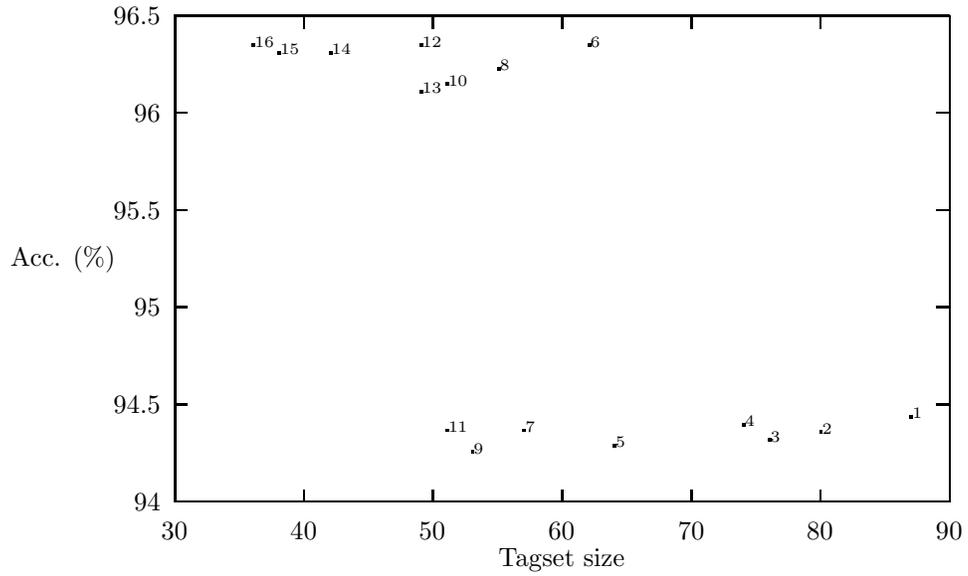
\begin{figure*}
\begin{center}
\item[]
\setlength{\unitlength}{0.240900pt}
\ifx\plotpoint\undefined\newsavebox{\plotpoint}\fi
\begin{picture}(1500,900)(0,0)
\font\gnuplot=cmr10 at 10pt
\gnuplot
\sbox{\plotpoint}{\rule[-0.200pt]{0.400pt}{0.400pt}}%
\put(220.0,113.0){\rule[-0.200pt]{4.818pt}{0.400pt}}
\put(198,113){\makebox(0,0)[r]{85}}
\put(1416.0,113.0){\rule[-0.200pt]{4.818pt}{0.400pt}}
\put(220.0,189.0){\rule[-0.200pt]{4.818pt}{0.400pt}}
\put(198,189){\makebox(0,0)[r]{85.5}}
\put(1416.0,189.0){\rule[-0.200pt]{4.818pt}{0.400pt}}
\put(220.0,266.0){\rule[-0.200pt]{4.818pt}{0.400pt}}
\put(198,266){\makebox(0,0)[r]{86}}
\put(1416.0,266.0){\rule[-0.200pt]{4.818pt}{0.400pt}}
\put(220.0,342.0){\rule[-0.200pt]{4.818pt}{0.400pt}}
\put(198,342){\makebox(0,0)[r]{86.5}}
\put(1416.0,342.0){\rule[-0.200pt]{4.818pt}{0.400pt}}
\put(220.0,419.0){\rule[-0.200pt]{4.818pt}{0.400pt}}
\put(198,419){\makebox(0,0)[r]{87}}
\put(1416.0,419.0){\rule[-0.200pt]{4.818pt}{0.400pt}}
\put(220.0,495.0){\rule[-0.200pt]{4.818pt}{0.400pt}}
\put(198,495){\makebox(0,0)[r]{87.5}}
\put(1416.0,495.0){\rule[-0.200pt]{4.818pt}{0.400pt}}
\put(220.0,571.0){\rule[-0.200pt]{4.818pt}{0.400pt}}
\put(198,571){\makebox(0,0)[r]{88}}
\put(1416.0,571.0){\rule[-0.200pt]{4.818pt}{0.400pt}}
\put(220.0,648.0){\rule[-0.200pt]{4.818pt}{0.400pt}}
\put(198,648){\makebox(0,0)[r]{88.5}}
\put(1416.0,648.0){\rule[-0.200pt]{4.818pt}{0.400pt}}
\put(220.0,724.0){\rule[-0.200pt]{4.818pt}{0.400pt}}
\put(198,724){\makebox(0,0)[r]{89}}
\put(1416.0,724.0){\rule[-0.200pt]{4.818pt}{0.400pt}}
\put(220.0,801.0){\rule[-0.200pt]{4.818pt}{0.400pt}}
\put(198,801){\makebox(0,0)[r]{89.5}}
\put(1416.0,801.0){\rule[-0.200pt]{4.818pt}{0.400pt}}
\put(220.0,877.0){\rule[-0.200pt]{4.818pt}{0.400pt}}
\put(198,877){\makebox(0,0)[r]{90}}
\put(1416.0,877.0){\rule[-0.200pt]{4.818pt}{0.400pt}}
\put(220.0,113.0){\rule[-0.200pt]{0.400pt}{4.818pt}}
\put(220,68){\makebox(0,0){90}}
\put(220.0,857.0){\rule[-0.200pt]{0.400pt}{4.818pt}}
\put(394.0,113.0){\rule[-0.200pt]{0.400pt}{4.818pt}}
\put(394,68){\makebox(0,0){100}}
\put(394.0,857.0){\rule[-0.200pt]{0.400pt}{4.818pt}}
\put(567.0,113.0){\rule[-0.200pt]{0.400pt}{4.818pt}}
\put(567,68){\makebox(0,0){110}}
\put(567.0,857.0){\rule[-0.200pt]{0.400pt}{4.818pt}}
\put(741.0,113.0){\rule[-0.200pt]{0.400pt}{4.818pt}}
\put(741,68){\makebox(0,0){120}}
\put(741.0,857.0){\rule[-0.200pt]{0.400pt}{4.818pt}}
\put(915.0,113.0){\rule[-0.200pt]{0.400pt}{4.818pt}}
\put(915,68){\makebox(0,0){130}}
\put(915.0,857.0){\rule[-0.200pt]{0.400pt}{4.818pt}}
\put(1089.0,113.0){\rule[-0.200pt]{0.400pt}{4.818pt}}
\put(1089,68){\makebox(0,0){140}}
\put(1089.0,857.0){\rule[-0.200pt]{0.400pt}{4.818pt}}
\put(1262.0,113.0){\rule[-0.200pt]{0.400pt}{4.818pt}}
\put(1262,68){\makebox(0,0){150}}
\put(1262.0,857.0){\rule[-0.200pt]{0.400pt}{4.818pt}}
\put(1436.0,113.0){\rule[-0.200pt]{0.400pt}{4.818pt}}
\put(1436,68){\makebox(0,0){160}}
\put(1436.0,857.0){\rule[-0.200pt]{0.400pt}{4.818pt}}
\put(220.0,113.0){\rule[-0.200pt]{292.934pt}{0.400pt}}
\put(1436.0,113.0){\rule[-0.200pt]{0.400pt}{184.048pt}}
\put(220.0,877.0){\rule[-0.200pt]{292.934pt}{0.400pt}}
\put(45,495){\makebox(0,0){Acc. (\%)}}
\put(828,23){\makebox(0,0){Tagset size}}
\put(220.0,113.0){\rule[-0.200pt]{0.400pt}{184.048pt}}
\put(1314,810){\rule{1pt}{1pt}{\tiny 1}}
\put(1262,765){\rule{1pt}{1pt}{\tiny 2}}
\put(1175,801){\rule{1pt}{1pt}{\tiny 3}}
\put(1089,775){\rule{1pt}{1pt}{\tiny 4}}
\put(1036,755){\rule{1pt}{1pt}{\tiny 5}}
\put(897,755){\rule{1pt}{1pt}{\tiny 6}}
\put(811,726){\rule{1pt}{1pt}{\tiny 7}}
\put(724,715){\rule{1pt}{1pt}{\tiny 8}}
\put(533,640){\rule{1pt}{1pt}{\tiny 9}}
\put(498,645){\rule{1pt}{1pt}{\tiny 10}}
\put(307,177){\rule{1pt}{1pt}{\tiny 11}}
\end{picture}
\caption{Results for English (no unknown words)}
\label{fig-eng1}
\end{center}
\end{figure*}
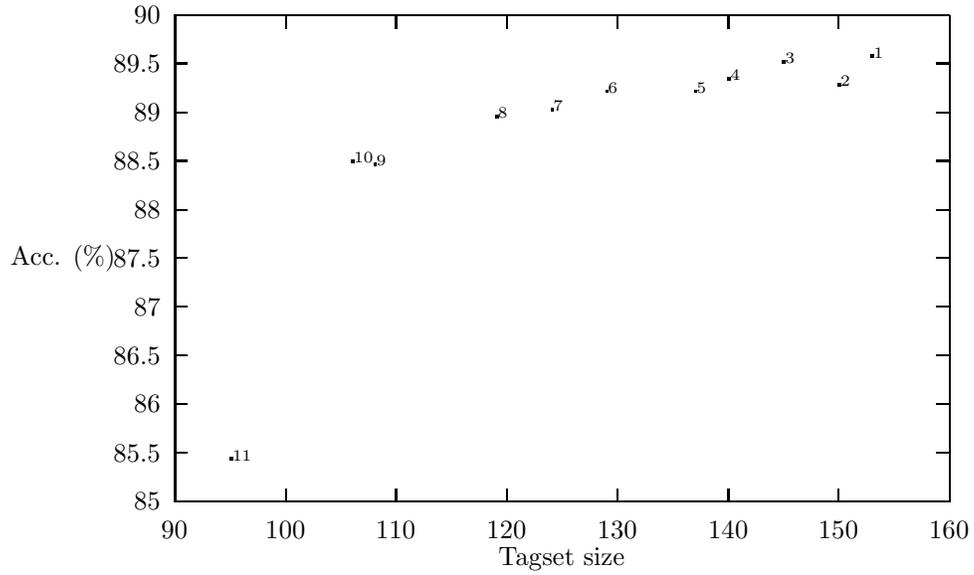

In the second experiment, the test corpora included ``unknown'' words, which
had not been seen during training, and for which the tagger hypothesises all
open-class tags. Two results are interesting to look at here: the accuracy on
the unknown words, and the accuracy on words which were ambiguous but were
found in the training corpus. The results, in outline, are:
\begin{description}
\item[Swedish] Similar results on known words to first experiment. For
unknown words, smaller tagsets give higher accuracy.
\item[French] For ambiguous words, the pattern and accuracy were similar to
first experiment. For unknown words, the pattern of accuracies was again
similar, with tagsets that do not include gender giving accuracies of
51\%--52\%, and those which do giving 45\%--46\%.
\item[English] Ambiguous words gave similar results to the first test. Unknown
words show a weak tendency to give higher accuracy on smaller tagsets.
\end{description}
Typical accuracies on ambiguous words were 90--92\%, 93--97\% and 83--88\% for
Swedish, French and English respectively, with the corresponding accuracies on
unknown words being 25--50\%, 45--52\% and
44--58\%. Table~\ref{tab2} lists the results, giving the
tagset size, the degree of ambiguity and the accuracies on known ambiguous and
unknown words. The ambiguous word accuracy is plotted in
figures~\ref{fig-swe2}--\ref{fig-eng2}.
\begin{table*}[p]
\begin{center}
\caption{Results for test with unknown words}
\begin{tabular}{|c|c|c|c|c|c|c|} \hline
Language & Index & Tagset & Size of & Degree of      & Ambiguous word & Unknown
word\\
         &       &        & tagset  & ambiguity (\%) & accuracy (\%)  &
accuracy (\%)\\ \hline
Swedish & 1 & GNDC & 194 & 52.60 & 91.09 & 23.42\\
        & 2 & GnDC & 170 & 50.56 & 91.62 & 26.28\\
        & 3 & GNDc & 167 & 52.59 & 91.01 & 24.17\\
        & 4 & GNdC & 162 & 52.48 & 90.77 & 28.48\\
        & 5 & gNDC & 152 & 52.57 & 91.19 & 29.33\\
        & 6 & GnDc & 147 & 50.55 & 91.51 & 26.48\\
        & 7 & GndC & 141 & 48.86 & 91.40 & 36.29\\
        & 8 & GNdc & 140 & 52.46 & 90.71 & 28.63\\
        & 9 & gNDc & 134 & 52.56 & 91.11 & 29.48\\
        & 10 & gNdC & 126 & 52.45 & 90.43 & 36.14\\
        & 11 & Gndc & 123 & 48.85 & 91.32 & 36.44\\
        & 12 & gnDC & 121 & 50.46 & 91.02 & 34.73\\
        & 13 & gNdc & 113 & 52.43 & 90.42 & 36.24\\
        & 14 & gnDc & 105 & 50.45 & 90.94 & 35.39\\
        & 15 & gndC & 96 & 48.75 & 91.09 & 48.00\\
        & 16 & gndc & 86 & 48.74 & 91.02 & 47.85\\\hline
French  & 1 & GNPV & 87 & 58.37 & 93.86 & 45.74\\
        & 2 & GNPv & 80 & 58.35 & 93.86 & 45.41\\
        & 3 & GNpV & 76 & 58.37 & 93.78 & 45.58\\
        & 4 & GNpv & 74 & 58.35 & 93.78 & 45.41\\
        & 5 & GnPV & 64 & 58.09 & 93.63 & 45.74\\
        & 6 & gNPV & 62 & 56.54 & 96.50 & 50.58\\
        & 7 & GnPv & 57 & 58.07 & 93.74 & 46.08\\
        & 8 & gNPv & 55 & 56.52 & 96.46 & 51.25\\
        & 9 & GnpV & 53 & 58.09 & 93.75 & 45.74\\
        & 10 & gNpV & 51 & 56.54 & 96.38 & 50.92\\
        & 11 & Gnpv & 51 & 58.07 & 93.78 & 46.24\\
        & 12 & gnPV & 49 & 56.09 & 96.26 & 50.92\\
        & 13 & gNpv & 49 & 56.62 & 96.34 & 50.75\\
        & 14 & gnPv & 42 & 56.08 & 96.26 & 52.45\\
        & 15 & gnpV & 38 & 56.09 & 96.26 & 52.25\\
        & 16 & gnpv & 36 & 56.08 & 96.34 & 52.59\\\hline
English & 1 & CAPNV & 153 & 55.65 & 87.57 & 46.49\\
        & 2 & CAPnv & 150 & 55.17 & 87.54 & 46.69\\
        & 3 & cAPNV & 145 & 55.60 & 87.46 & 46.29\\
        & 4 & CAPNv & 140 & 55.65 & 87.52 & 46.09\\
        & 5 & CAPnV & 137 & 55.65 & 87.62 & 51.70\\
        & 6 & CaPNV & 129 & 55.65 & 87.43 & 46.29\\
        & 7 & CAPnv & 124 & 55.65 & 87.48 & 51.70\\
        & 8 & capNV & 119 & 55.13 & 87.38 & 46.49\\
        & 9 & capnV & 108 & 55.00 & 83.66 & 56.51\\
        & 10 & capNv & 106 & 55.13 & 83.66 & 44.29\\
        & 11 & capnv & 95 & 55.00 & 83.56 & 55.11\\\hline
\end{tabular}
\label{tab2}
\end{center}
\end{table*}
\begin{figure*}
\begin{center}
\item[]
\setlength{\unitlength}{0.240900pt}
\ifx\plotpoint\undefined\newsavebox{\plotpoint}\fi
\begin{picture}(1500,900)(0,0)
\font\gnuplot=cmr10 at 10pt
\gnuplot
\sbox{\plotpoint}{\rule[-0.200pt]{0.400pt}{0.400pt}}%
\put(220.0,113.0){\rule[-0.200pt]{4.818pt}{0.400pt}}
\put(198,113){\makebox(0,0)[r]{90.4}}
\put(1416.0,113.0){\rule[-0.200pt]{4.818pt}{0.400pt}}
\put(220.0,222.0){\rule[-0.200pt]{4.818pt}{0.400pt}}
\put(198,222){\makebox(0,0)[r]{90.6}}
\put(1416.0,222.0){\rule[-0.200pt]{4.818pt}{0.400pt}}
\put(220.0,331.0){\rule[-0.200pt]{4.818pt}{0.400pt}}
\put(198,331){\makebox(0,0)[r]{90.8}}
\put(1416.0,331.0){\rule[-0.200pt]{4.818pt}{0.400pt}}
\put(220.0,440.0){\rule[-0.200pt]{4.818pt}{0.400pt}}
\put(198,440){\makebox(0,0)[r]{91}}
\put(1416.0,440.0){\rule[-0.200pt]{4.818pt}{0.400pt}}
\put(220.0,550.0){\rule[-0.200pt]{4.818pt}{0.400pt}}
\put(198,550){\makebox(0,0)[r]{91.2}}
\put(1416.0,550.0){\rule[-0.200pt]{4.818pt}{0.400pt}}
\put(220.0,659.0){\rule[-0.200pt]{4.818pt}{0.400pt}}
\put(198,659){\makebox(0,0)[r]{91.4}}
\put(1416.0,659.0){\rule[-0.200pt]{4.818pt}{0.400pt}}
\put(220.0,768.0){\rule[-0.200pt]{4.818pt}{0.400pt}}
\put(198,768){\makebox(0,0)[r]{91.6}}
\put(1416.0,768.0){\rule[-0.200pt]{4.818pt}{0.400pt}}
\put(220.0,877.0){\rule[-0.200pt]{4.818pt}{0.400pt}}
\put(198,877){\makebox(0,0)[r]{91.8}}
\put(1416.0,877.0){\rule[-0.200pt]{4.818pt}{0.400pt}}
\put(220.0,113.0){\rule[-0.200pt]{0.400pt}{4.818pt}}
\put(220,68){\makebox(0,0){80}}
\put(220.0,857.0){\rule[-0.200pt]{0.400pt}{4.818pt}}
\put(423.0,113.0){\rule[-0.200pt]{0.400pt}{4.818pt}}
\put(423,68){\makebox(0,0){100}}
\put(423.0,857.0){\rule[-0.200pt]{0.400pt}{4.818pt}}
\put(625.0,113.0){\rule[-0.200pt]{0.400pt}{4.818pt}}
\put(625,68){\makebox(0,0){120}}
\put(625.0,857.0){\rule[-0.200pt]{0.400pt}{4.818pt}}
\put(828.0,113.0){\rule[-0.200pt]{0.400pt}{4.818pt}}
\put(828,68){\makebox(0,0){140}}
\put(828.0,857.0){\rule[-0.200pt]{0.400pt}{4.818pt}}
\put(1031.0,113.0){\rule[-0.200pt]{0.400pt}{4.818pt}}
\put(1031,68){\makebox(0,0){160}}
\put(1031.0,857.0){\rule[-0.200pt]{0.400pt}{4.818pt}}
\put(1233.0,113.0){\rule[-0.200pt]{0.400pt}{4.818pt}}
\put(1233,68){\makebox(0,0){180}}
\put(1233.0,857.0){\rule[-0.200pt]{0.400pt}{4.818pt}}
\put(1436.0,113.0){\rule[-0.200pt]{0.400pt}{4.818pt}}
\put(1436,68){\makebox(0,0){200}}
\put(1436.0,857.0){\rule[-0.200pt]{0.400pt}{4.818pt}}
\put(220.0,113.0){\rule[-0.200pt]{292.934pt}{0.400pt}}
\put(1436.0,113.0){\rule[-0.200pt]{0.400pt}{184.048pt}}
\put(220.0,877.0){\rule[-0.200pt]{292.934pt}{0.400pt}}
\put(45,495){\makebox(0,0){Acc. (\%)}}
\put(828,23){\makebox(0,0){Tagset size}}
\put(220.0,113.0){\rule[-0.200pt]{0.400pt}{184.048pt}}
\put(1375,490){\rule{1pt}{1pt}{\tiny 1}}
\put(1132,779){\rule{1pt}{1pt}{\tiny 2}}
\put(1102,446){\rule{1pt}{1pt}{\tiny 3}}
\put(1051,315){\rule{1pt}{1pt}{\tiny 4}}
\put(950,544){\rule{1pt}{1pt}{\tiny 5}}
\put(899,719){\rule{1pt}{1pt}{\tiny 6}}
\put(838,659){\rule{1pt}{1pt}{\tiny 7}}
\put(828,282){\rule{1pt}{1pt}{\tiny 8}}
\put(767,500){\rule{1pt}{1pt}{\tiny 9}}
\put(686,129){\rule{1pt}{1pt}{\tiny 10}}
\put(656,615){\rule{1pt}{1pt}{\tiny 11}}
\put(635,451){\rule{1pt}{1pt}{\tiny 12}}
\put(554,124){\rule{1pt}{1pt}{\tiny 13}}
\put(473,408){\rule{1pt}{1pt}{\tiny 14}}
\put(382,490){\rule{1pt}{1pt}{\tiny 15}}
\put(281,451){\rule{1pt}{1pt}{\tiny 16}}
\end{picture}
\caption{Results for Swedish (with unknown words)}
\label{fig-swe2}
\end{center}
\end{figure*}
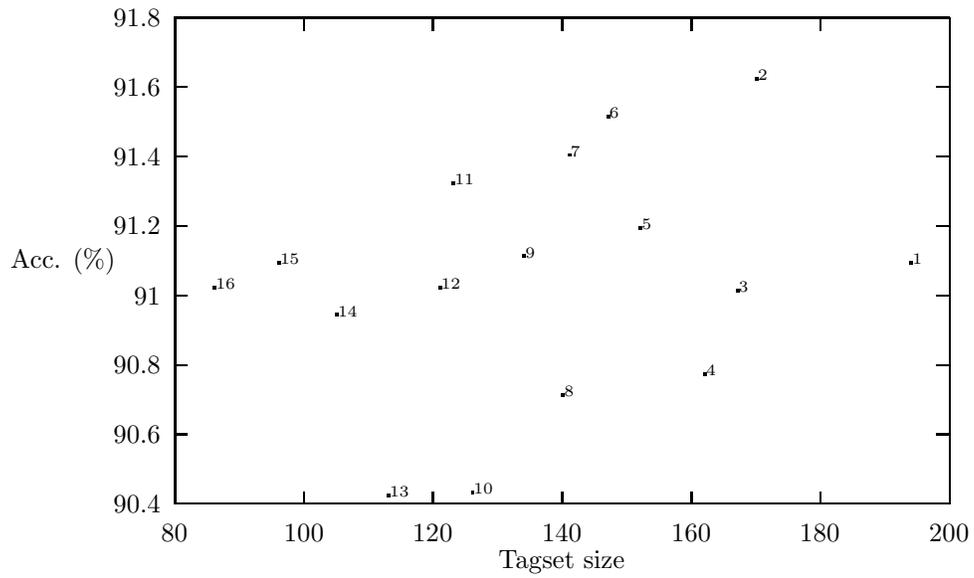
\begin{figure*}
\begin{center}
\item[]
\setlength{\unitlength}{0.240900pt}
\ifx\plotpoint\undefined\newsavebox{\plotpoint}\fi
\begin{picture}(1500,900)(0,0)
\font\gnuplot=cmr10 at 10pt
\gnuplot
\sbox{\plotpoint}{\rule[-0.200pt]{0.400pt}{0.400pt}}%
\put(220.0,215.0){\rule[-0.200pt]{4.818pt}{0.400pt}}
\put(198,215){\makebox(0,0)[r]{94}}
\put(1416.0,215.0){\rule[-0.200pt]{4.818pt}{0.400pt}}
\put(220.0,342.0){\rule[-0.200pt]{4.818pt}{0.400pt}}
\put(198,342){\makebox(0,0)[r]{94.5}}
\put(1416.0,342.0){\rule[-0.200pt]{4.818pt}{0.400pt}}
\put(220.0,470.0){\rule[-0.200pt]{4.818pt}{0.400pt}}
\put(198,470){\makebox(0,0)[r]{95}}
\put(1416.0,470.0){\rule[-0.200pt]{4.818pt}{0.400pt}}
\put(220.0,597.0){\rule[-0.200pt]{4.818pt}{0.400pt}}
\put(198,597){\makebox(0,0)[r]{95.5}}
\put(1416.0,597.0){\rule[-0.200pt]{4.818pt}{0.400pt}}
\put(220.0,724.0){\rule[-0.200pt]{4.818pt}{0.400pt}}
\put(198,724){\makebox(0,0)[r]{96}}
\put(1416.0,724.0){\rule[-0.200pt]{4.818pt}{0.400pt}}
\put(220.0,852.0){\rule[-0.200pt]{4.818pt}{0.400pt}}
\put(198,852){\makebox(0,0)[r]{96.5}}
\put(1416.0,852.0){\rule[-0.200pt]{4.818pt}{0.400pt}}
\put(220.0,113.0){\rule[-0.200pt]{0.400pt}{4.818pt}}
\put(220,68){\makebox(0,0){30}}
\put(220.0,857.0){\rule[-0.200pt]{0.400pt}{4.818pt}}
\put(423.0,113.0){\rule[-0.200pt]{0.400pt}{4.818pt}}
\put(423,68){\makebox(0,0){40}}
\put(423.0,857.0){\rule[-0.200pt]{0.400pt}{4.818pt}}
\put(625.0,113.0){\rule[-0.200pt]{0.400pt}{4.818pt}}
\put(625,68){\makebox(0,0){50}}
\put(625.0,857.0){\rule[-0.200pt]{0.400pt}{4.818pt}}
\put(828.0,113.0){\rule[-0.200pt]{0.400pt}{4.818pt}}
\put(828,68){\makebox(0,0){60}}
\put(828.0,857.0){\rule[-0.200pt]{0.400pt}{4.818pt}}
\put(1031.0,113.0){\rule[-0.200pt]{0.400pt}{4.818pt}}
\put(1031,68){\makebox(0,0){70}}
\put(1031.0,857.0){\rule[-0.200pt]{0.400pt}{4.818pt}}
\put(1233.0,113.0){\rule[-0.200pt]{0.400pt}{4.818pt}}
\put(1233,68){\makebox(0,0){80}}
\put(1233.0,857.0){\rule[-0.200pt]{0.400pt}{4.818pt}}
\put(1436.0,113.0){\rule[-0.200pt]{0.400pt}{4.818pt}}
\put(1436,68){\makebox(0,0){90}}
\put(1436.0,857.0){\rule[-0.200pt]{0.400pt}{4.818pt}}
\put(220.0,113.0){\rule[-0.200pt]{292.934pt}{0.400pt}}
\put(1436.0,113.0){\rule[-0.200pt]{0.400pt}{184.048pt}}
\put(220.0,877.0){\rule[-0.200pt]{292.934pt}{0.400pt}}
\put(45,495){\makebox(0,0){Acc. (\%)}}
\put(828,23){\makebox(0,0){Tagset size}}
\put(220.0,113.0){\rule[-0.200pt]{0.400pt}{184.048pt}}
\put(1375,179){\rule{1pt}{1pt}{\tiny 1}}
\put(1233,179){\rule{1pt}{1pt}{\tiny 2}}
\put(1152,159){\rule{1pt}{1pt}{\tiny 3}}
\put(1112,159){\rule{1pt}{1pt}{\tiny 4}}
\put(909,121){\rule{1pt}{1pt}{\tiny 5}}
\put(869,852){\rule{1pt}{1pt}{\tiny 6}}
\put(767,149){\rule{1pt}{1pt}{\tiny 7}}
\put(727,841){\rule{1pt}{1pt}{\tiny 8}}
\put(686,151){\rule{1pt}{1pt}{\tiny 9}}
\put(646,821){\rule{1pt}{1pt}{\tiny 10}}
\put(646,159){\rule{1pt}{1pt}{\tiny 11}}
\put(605,790){\rule{1pt}{1pt}{\tiny 12}}
\put(605,811){\rule{1pt}{1pt}{\tiny 13}}
\put(463,790){\rule{1pt}{1pt}{\tiny 14}}
\put(382,790){\rule{1pt}{1pt}{\tiny 15}}
\put(342,811){\rule{1pt}{1pt}{\tiny 16}}
\end{picture}
\caption{Results for French (with unknown words)}
\label{fig-fre2}
\end{center}
\end{figure*}
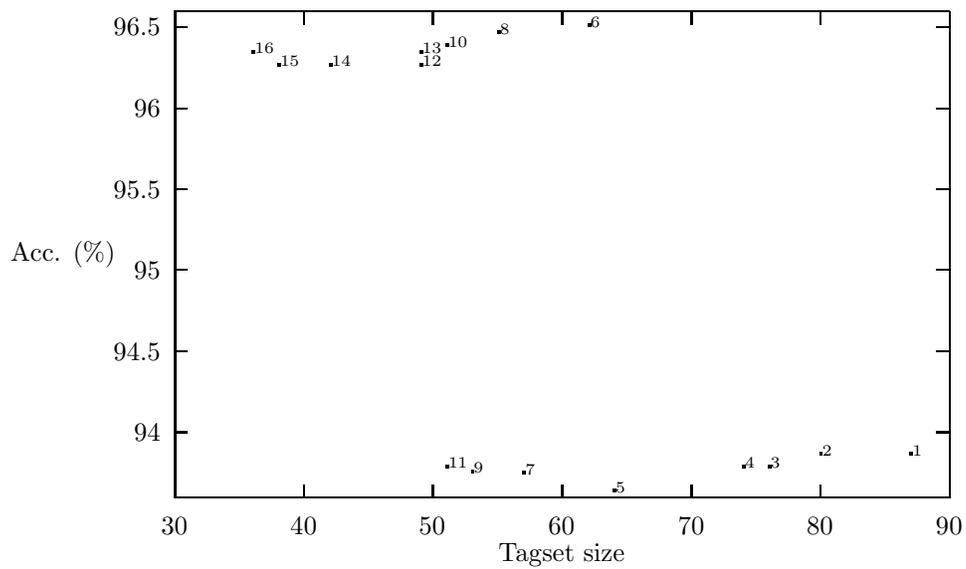
\begin{figure*}
\begin{center}
\item[]
\setlength{\unitlength}{0.240900pt}
\ifx\plotpoint\undefined\newsavebox{\plotpoint}\fi
\begin{picture}(1500,900)(0,0)
\font\gnuplot=cmr10 at 10pt
\gnuplot
\sbox{\plotpoint}{\rule[-0.200pt]{0.400pt}{0.400pt}}%
\put(220.0,113.0){\rule[-0.200pt]{4.818pt}{0.400pt}}
\put(198,113){\makebox(0,0)[r]{83.5}}
\put(1416.0,113.0){\rule[-0.200pt]{4.818pt}{0.400pt}}
\put(220.0,198.0){\rule[-0.200pt]{4.818pt}{0.400pt}}
\put(198,198){\makebox(0,0)[r]{84}}
\put(1416.0,198.0){\rule[-0.200pt]{4.818pt}{0.400pt}}
\put(220.0,283.0){\rule[-0.200pt]{4.818pt}{0.400pt}}
\put(198,283){\makebox(0,0)[r]{84.5}}
\put(1416.0,283.0){\rule[-0.200pt]{4.818pt}{0.400pt}}
\put(220.0,368.0){\rule[-0.200pt]{4.818pt}{0.400pt}}
\put(198,368){\makebox(0,0)[r]{85}}
\put(1416.0,368.0){\rule[-0.200pt]{4.818pt}{0.400pt}}
\put(220.0,453.0){\rule[-0.200pt]{4.818pt}{0.400pt}}
\put(198,453){\makebox(0,0)[r]{85.5}}
\put(1416.0,453.0){\rule[-0.200pt]{4.818pt}{0.400pt}}
\put(220.0,537.0){\rule[-0.200pt]{4.818pt}{0.400pt}}
\put(198,537){\makebox(0,0)[r]{86}}
\put(1416.0,537.0){\rule[-0.200pt]{4.818pt}{0.400pt}}
\put(220.0,622.0){\rule[-0.200pt]{4.818pt}{0.400pt}}
\put(198,622){\makebox(0,0)[r]{86.5}}
\put(1416.0,622.0){\rule[-0.200pt]{4.818pt}{0.400pt}}
\put(220.0,707.0){\rule[-0.200pt]{4.818pt}{0.400pt}}
\put(198,707){\makebox(0,0)[r]{87}}
\put(1416.0,707.0){\rule[-0.200pt]{4.818pt}{0.400pt}}
\put(220.0,792.0){\rule[-0.200pt]{4.818pt}{0.400pt}}
\put(198,792){\makebox(0,0)[r]{87.5}}
\put(1416.0,792.0){\rule[-0.200pt]{4.818pt}{0.400pt}}
\put(220.0,877.0){\rule[-0.200pt]{4.818pt}{0.400pt}}
\put(198,877){\makebox(0,0)[r]{88}}
\put(1416.0,877.0){\rule[-0.200pt]{4.818pt}{0.400pt}}
\put(220.0,113.0){\rule[-0.200pt]{0.400pt}{4.818pt}}
\put(220,68){\makebox(0,0){90}}
\put(220.0,857.0){\rule[-0.200pt]{0.400pt}{4.818pt}}
\put(394.0,113.0){\rule[-0.200pt]{0.400pt}{4.818pt}}
\put(394,68){\makebox(0,0){100}}
\put(394.0,857.0){\rule[-0.200pt]{0.400pt}{4.818pt}}
\put(567.0,113.0){\rule[-0.200pt]{0.400pt}{4.818pt}}
\put(567,68){\makebox(0,0){110}}
\put(567.0,857.0){\rule[-0.200pt]{0.400pt}{4.818pt}}
\put(741.0,113.0){\rule[-0.200pt]{0.400pt}{4.818pt}}
\put(741,68){\makebox(0,0){120}}
\put(741.0,857.0){\rule[-0.200pt]{0.400pt}{4.818pt}}
\put(915.0,113.0){\rule[-0.200pt]{0.400pt}{4.818pt}}
\put(915,68){\makebox(0,0){130}}
\put(915.0,857.0){\rule[-0.200pt]{0.400pt}{4.818pt}}
\put(1089.0,113.0){\rule[-0.200pt]{0.400pt}{4.818pt}}
\put(1089,68){\makebox(0,0){140}}
\put(1089.0,857.0){\rule[-0.200pt]{0.400pt}{4.818pt}}
\put(1262.0,113.0){\rule[-0.200pt]{0.400pt}{4.818pt}}
\put(1262,68){\makebox(0,0){150}}
\put(1262.0,857.0){\rule[-0.200pt]{0.400pt}{4.818pt}}
\put(1436.0,113.0){\rule[-0.200pt]{0.400pt}{4.818pt}}
\put(1436,68){\makebox(0,0){160}}
\put(1436.0,857.0){\rule[-0.200pt]{0.400pt}{4.818pt}}
\put(220.0,113.0){\rule[-0.200pt]{292.934pt}{0.400pt}}
\put(1436.0,113.0){\rule[-0.200pt]{0.400pt}{184.048pt}}
\put(220.0,877.0){\rule[-0.200pt]{292.934pt}{0.400pt}}
\put(45,495){\makebox(0,0){Acc. (\%)}}
\put(828,23){\makebox(0,0){Tagset size}}
\put(220.0,113.0){\rule[-0.200pt]{0.400pt}{184.048pt}}
\put(1314,804){\rule{1pt}{1pt}{\tiny 1}}
\put(1262,799){\rule{1pt}{1pt}{\tiny 2}}
\put(1175,785){\rule{1pt}{1pt}{\tiny 3}}
\put(1089,796){\rule{1pt}{1pt}{\tiny 4}}
\put(1036,812){\rule{1pt}{1pt}{\tiny 5}}
\put(897,780){\rule{1pt}{1pt}{\tiny 6}}
\put(811,789){\rule{1pt}{1pt}{\tiny 7}}
\put(724,772){\rule{1pt}{1pt}{\tiny 8}}
\put(533,140){\rule{1pt}{1pt}{\tiny 9}}
\put(498,140){\rule{1pt}{1pt}{\tiny 10}}
\put(307,123){\rule{1pt}{1pt}{\tiny 11}}
\end{picture}
\caption{Results for English (with unknown words)}
\label{fig-eng2}
\end{center}
\end{figure*}
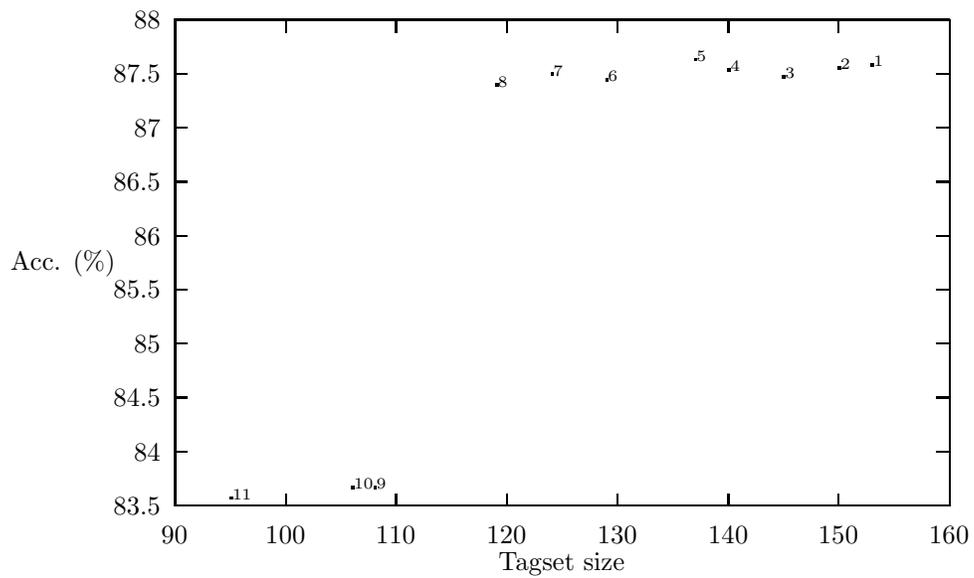

What seems to come out from these results is that there is not a consistent
relationship between the size of the tagsets and the tagging accuracy. The
most common pattern was for a larger tagset to give higher accuracy, but there
were notable exceptions in French (where gender marking was the key factor),
in Swedish unknown words (which show the reverse trend) and in English unknown
words (which show no very clear trend at all). This seems to fit quite well
with the difficulties that were suggested above in reasoning about the effect
of tagset size. The main conclusion of this paper is therefore that the
knowledge engineering component of setting up a tagger should concentrate on
optimising the tagset for external criteria, and that the internal criterion
of tagset size does not show sufficient generality to be taken into account
without prior knowledge of properties of the language. Perhaps this is not too
surprising, but it is useful to have an experimental confirmation that the
linguistics matters rather than the engineering.

\section{Unknown words}
One final observation about the experiments: the accuracy on unknown words was
very low in all of the tests, and was particularly bad in Swedish. The
tagger used in the experiments took a very simple-minded approach to unknown
words. An alternative that is often used is to limit the possible tags using a
simple morphological analysis or some other examination of the surface form of
the word. For example, in a variant of the English tagger which was not used
in these experiments, a module which reduces the range of possible tags based
on testing for only seven surface characteristics such as capitalisation and
word endings improved the unknown word accuracy by 15-20\%.

The results above show that if it were not for unknown words, there might be
some argument for favouring larger tagsets, since they have some tendency to
give a higher accuracy. A tentative experiment on the contribution of using
morphological or surface analysis in French and Swedish was therefore carried
out. Firstly, in both languages, the unknown words from the second experiment
were looked up in the lexicon trained from the full corpus to see what tags
they might have. For Swedish, 96\% of the unknown words came from inflected
classes, and had a single tag; for French the figure was about 60\%. In both
cases, very few of the unknown words (less than 1\%) had more than one
tag. This provides some hope that an inflectional analysis might should help
considerably with unknown words\footnote{Although the figures here are likely
to represent a best case, given how little of the corpora was held out.}. For
confirmation, the list of French unknown words was given to a French
grammarian, who predicted that it would be possible to make a good guess at
the correct tag from the morphology for around 70\% of the words, and could
narrow down the possible tags to two or three for about a further
25\%. However, further research is needed to determine how realistic these
estimates turn out to be.

\section{Conclusion}

We have shown how a simple experiment in changing the tagset shows that the
relationship between tagset size and accuracy is a weak one and is not
consistent against languages. This seems to go against the ``folklore'' of the
tagging community, where smaller tagsets are often held to be better for
obtaining good accuracy. I have suggested that what is important is to choose
the tagset required for the application, rather than to optimise it for the
tagger. A follow-up to this work might be to apply similar tests in other
languages to provide a further confirmation of the results, and to see if
language families which similar characteristics can be identified. A further
conclusion might be that when a corpus is being tagged by hand, a large tagset
should be used, since it can always be reduced to a smaller one if the
application demands it. Perhaps the major factor we have to set against this
is the danger of introducing more human errors into the manual tagging
process, by increasing the cognitive load on the human annotators.

{}


\begin{thebibliography}{}

\bibitem[\protect\citename{Cutting et al.}1992]{Cutting:pos}
Doug Cutting, Julian Kupiec, Jan Pedersen, and Penelope Sibun (1992).
\newblock A Practical Part-of-Speech Tagger.
\newblock In {\it Third Conference on Applied Natural Language Processing.
  Proceedings of the Conference. Trento, Italy}, pages~133--140, Association
  for Computational Linguistics.

\bibitem[\protect\citename{Elworthy}1994]{Elworthy:help}
David Elworthy (1994).
\newblock Does Baum-Welch Re-estimation Help Taggers?
\newblock In {\it Fourth Conference on Applied Natural Language Processing.
  Proceedings of the Conference. Stuttgart, Germany}, pages~53--58, Association
  for Computational Linguistics.

\bibitem[\protect\citename{Francis and Ku\v{c}era}1992]{Francis:brown}
W.~N. Francis and F.~Ku\v{c}era (1992).
\newblock {\it Frequency Analysis of {E}nglish Usage}.
\newblock Houghton Mifflin.

\bibitem[\protect\citename{Garside et al.}1987]{Garside:LAE}
Roger Garside, Geoffrey Leech, and Geoffrey Sampson (1987).
\newblock {\it The Computational Analysis of {E}nglish: A Corpus-based
  Approach}.
\newblock Longman, London.

\bibitem[\protect\citename{Macklovitch}1992]{Macklovitch:falter}
Elliott Macklovitch (1992).
\newblock Where the Tagger Falters.
\newblock In {\it Proceedings of the Fourth Conference on Theoretical and
  Methodological Issues in Machine Translation}, pages~113--126.

\bibitem[\protect\citename{Marcus et al.}1993]{Marcus:penn}
Mitchell~P. Marcus, Beatrice Santorini, and Mary~Ann Marcinkiewicz (1993).
\newblock Building a Large Annotated Corpus of {E}nglish: The {Penn} Treebank.
\newblock {\it Computational Linguistics}, 19(2):313--330.

\bibitem[\protect\citename{McEnery et al.}1994]{McEnery:parallel}
A.~M. McEnery, M.~P. Oakes, R.~Garside, J.~Hutchinson, and G.~N. Leech (1994).
\newblock The Exploitation of Parallel Corpora in Projects ET10/63 and CRATER.
\newblock In {\it International Conference on New Methods in Language
  Processing. Proceedings of the Conference}, pages~108--115, Centre for
  Computational Linguistics, UMIST.

\bibitem[\protect\citename{Merialdo}1994]{Merialdo:tagging}
Bernard Merialdo (1994).
\newblock Tagging English Text with a Probabilistic Model.
\newblock {\it Computational Linguistics}, 20(2):155--171.

\end{thebibliography}
\end{document}